\pdfoutput=1

\documentclass[11pt]{article}

\usepackage[]{acl}

\usepackage{times}
\usepackage{latexsym}

\usepackage[T1]{fontenc}

\usepackage[utf8]{inputenc}

\usepackage{microtype}
\usepackage{graphicx}
\usepackage{subfigure}
\usepackage{booktabs}
\usepackage{multirow}
\usepackage{bbding}
\usepackage{pifont}
\usepackage[nointegrals]{wasysym}
\usepackage{amssymb}
\usepackage{bm}
\usepackage{amsmath}
\usepackage{subfigure}
\usepackage{parskip}
\title{CREATE: A Benchmark for Chinese Short Video Retrieval and \\ Title Generation}

\author{Ziqi Zhang, Yuxin Chen, Zongyang Ma, Chunfeng Yuan, Bing Li, Weiming Hu \\ 
NLPR, Institute of Automation, Chinese Academy of Sciences \\
School of Artificial Intelligence, University of Chinese Academy of Sciences\\
\texttt{{zhangziqi2017,mazongyang2020}@ia.ac.cn,chenyux53@163.com}
        \AND
        Zhongang Qi, Ying Shan \\ Applied Research Center (ARC), Tencent PCG\\
        \texttt{{zhongangqi,yingsshan}@tencent.com}}

\begin{document}
\maketitle

\begin{abstract}
Previous works of video captioning aim to objectively describe the video's actual content, lack of subjective and attractive expression, limiting its practical application scenarios. Video titling is intended to achieve this goal, but there is a lack of a proper benchmark.
In this paper, we propose CREATE, the first large-scale \textbf{C}hinese sho\textbf{R}t vid\textbf{E}o retriev\textbf{A}l and \textbf{T}itle g\textbf{E}neration benchmark, to facilitate research and application in video titling and video retrieval in Chinese.
CREATE consists of a high-quality labeled 210K dataset and two large-scale 3M/10M pre-training datasets, covering 51 categories, 50K+ tags, 537K manually annotated titles and captions, and 10M+ short videos. 
Based on CREATE, we propose a novel model ALWIG which combines video retrieval and video titling tasks to achieve the purpose of multi-modal \textbf{AL}ignment \textbf{WI}th \textbf{G}eneration with the help of video tags and GPT pre-trained model.
CREATE opens new directions for facilitating future research and applications on video titling and video retrieval in the field of Chinese short videos.
\end{abstract}

\section{Introduction}

The video captioning task is gaining increasing attention in the vision and language communities. Although many research efforts have been made in both advanced algorithms \cite{scene, Attribute, Object, Object-Relation} as well as large-scale benchmarks in a variety of domains \cite{TVR,Youcook2, video-news}, there is little practical application being landed around video captioning.

\begin{figure}
    \centering
    \includegraphics[width=1\linewidth]{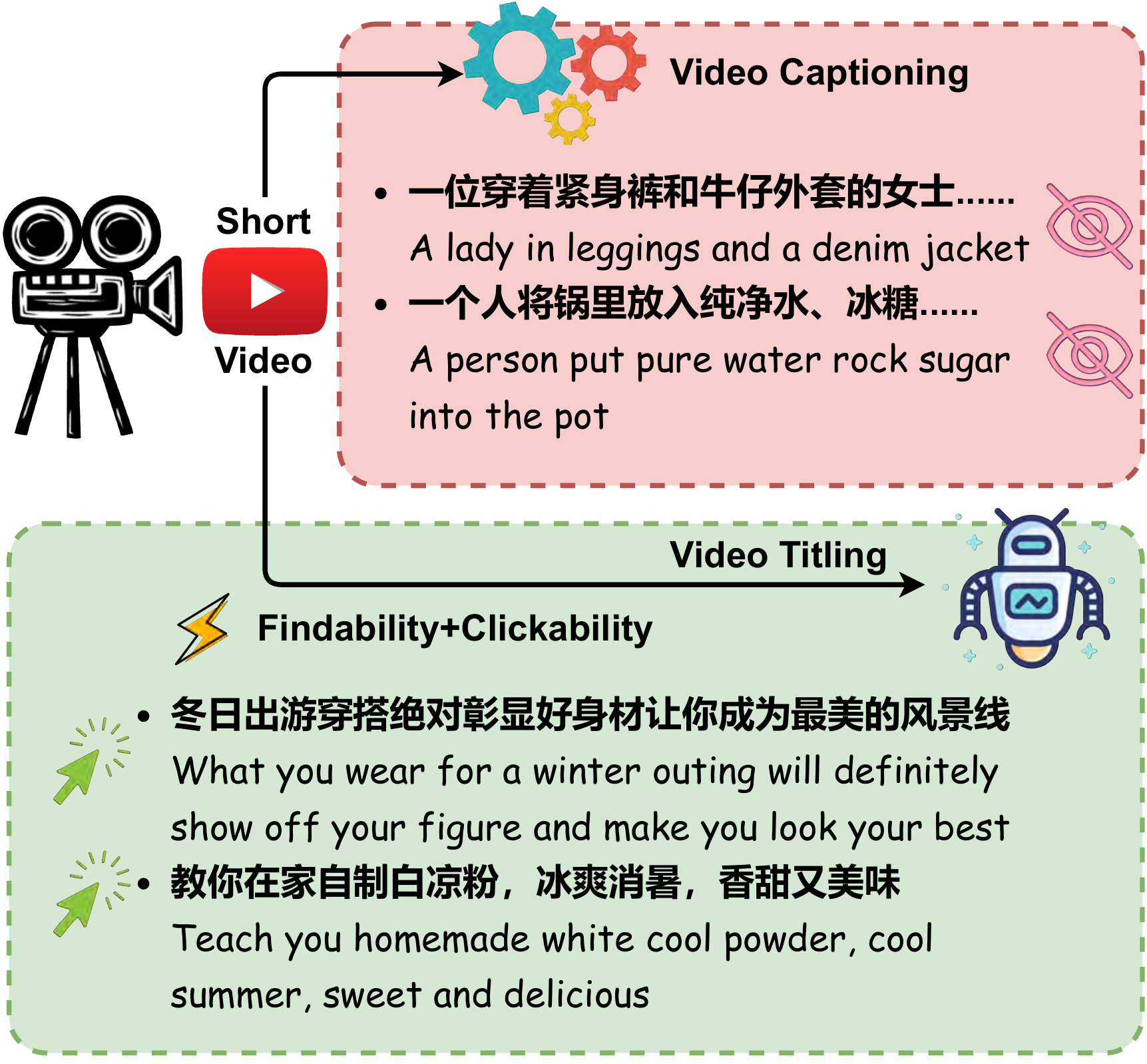}
    \caption{\textbf{Video captioning \textit{vs.} video titling.} Basically, captions are short factual summaries, while titles are what would be displayed to users to encourage them to watch the video.}
    \label{fig:caption-title}
\end{figure}

Captioning intends to give an image or a video clip an appropriate title in the newspaper or other social media. However, due to the influence of existing datasets, the video captioning task has developed into an objective description of the actual content of the video without subjective factors, which is not consistent with the practical application, as shown in Figure~\ref{fig:caption-title}. The primary reason is the gap between existing benchmarks and application scenarios.

\begin{figure*}
    \centering
    \includegraphics[width=1\linewidth]{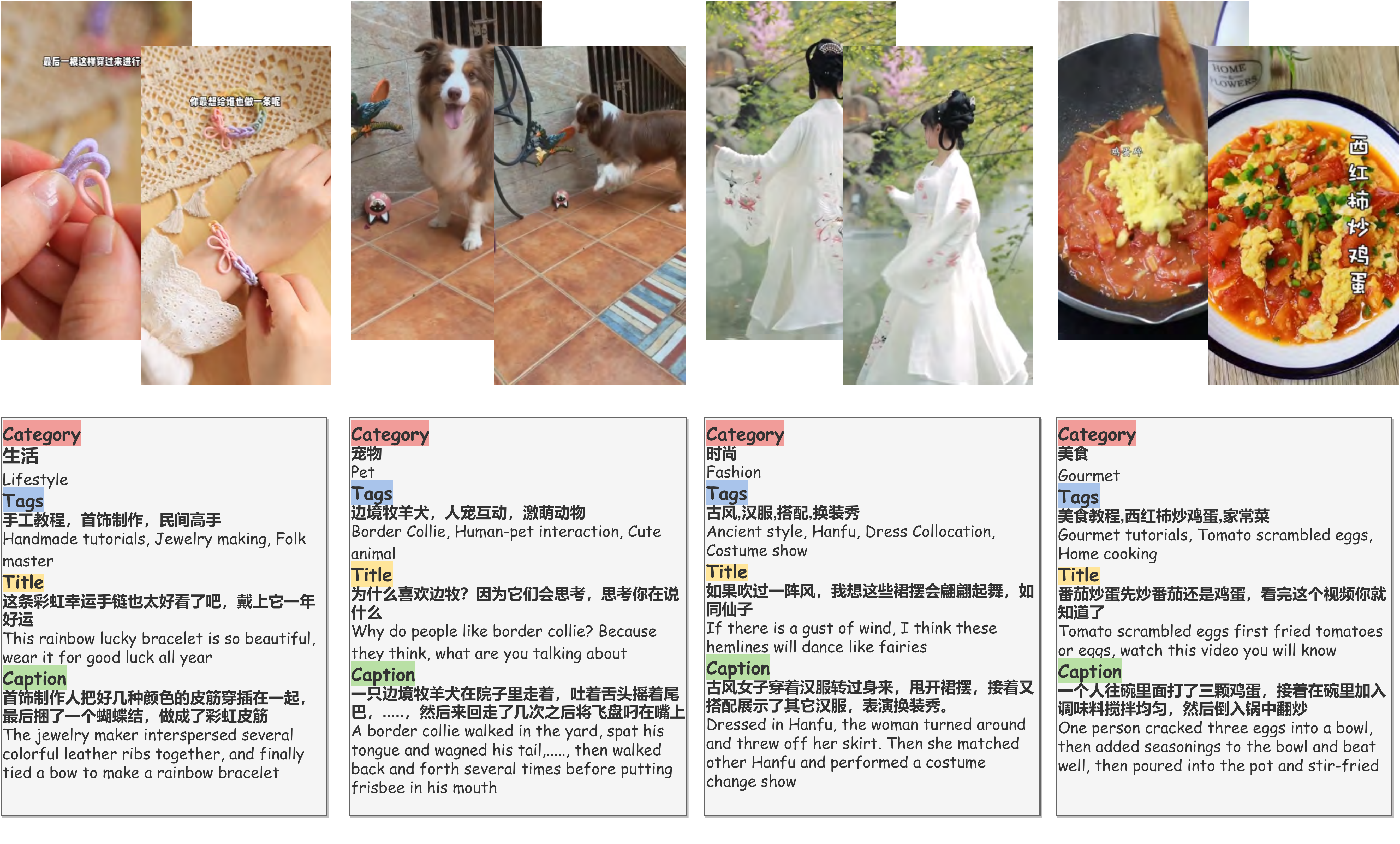}
    \caption{\textbf{A glance at the annotations in our CREATE benchmark.} It covers 51 categories such as Lifestyle, Pet, Fashion, Gourmet, \emph{etc.}, as well as 50K+ fine-grained tags. Each short video is annotated by an objective caption and a catchy title in an actual scenario.}
    \label{fig:show-dataset}
\end{figure*}

Most successful video starts with a good video title. The two components needed for crafting the best video titles are \textit{findability} and \textit{clickability}. The former requires to state the primary viewpoints of the video to facilitate the text-based search, while the latter expects to add catchy expressions to hook more viewers. Therefore, an automatic video title generator with both abilities can help junior creators solve this tricky problem. 

Chinese short videos play an important role in the global market, but the study of Chinese corpus is not enough.  The alignment of Chinese corpus with visual content is vital for the comprehension and creation of Chinese short videos. 
Therefore, it is necessary to pave the path for research and applications around Chinese short video titling and retrieval by establishing a new large-scale benchmark covering video titles and captions in Chinese.

To this end, we create the first \textbf{C}hinese sho\textbf{R}t vid\textbf{E}o retriev\textbf{A}l and \textbf{T}itle g\textbf{E}neration benchmark called CREATE. 
It contains two parts, the fine-labeled CREATE-210K and weak-labeled CREATE3M/10M. The CREATE-210K consists of 216K carefully collected short videos covering 51 categories and 15.5K tags, as illustrated in Figure~\ref{fig:show-dataset}. Each video is equipped with a high-quality title and caption to serve tasks such as video retrieval, tagging, titling and captioning. 
The CREATE-3M/10M are two large-scale datasets containing approximately 3M/10M videos with original titles and 53K tags. It can be used to learn vision and language alignment in the setting of weak-supervised learning through pre-training tasks.

It is worth noting that the number of videos in our CREATE-210K is 5.23 times that of VATEX \cite{VATEX}, the largest Chinese caption dataset, and the number of annotations is 2.98 times that of T-VTD, the largest e-commerce title dataset, as shown in Table \ref{tab:compare-dataset}. Large-scale pre-training datasets CREATE-3M/10M provide more diverse training methods. In addition, the annotations are encouraged to make full reference to audio, character, speech, and other fine-grained entities, such as celebrities, locations, and popular objects in the video, to enhance the semantic representation of the model in future research.

Based on this benchmark, we propose a novel vision and language model (VLM) called ALWIG, which combines video retrieval and video titling tasks to achieve the purpose of multi-modal \textbf{AL}ignment \textbf{WI}th \textbf{G}eneration. Specifically, we utilize the tag-driven module to achieve the alignment between visual and text. 
We take advantage of the powerful generative capability of GPT  \cite{gpt2} as the decoder for the textual generation. Meanwhile, we set up two popular pre-trained VLMs, \emph{i.e.}, OSCAR \cite{oscar} and UniVL \cite{univl}, as baseline models. The experimental results highlight the benefits of our method.

The main contribution of this paper is three-fold:

\begin{itemize}
    \item 
    We establish the first large-scale benchmark CREATE for Chinese short video titling and retrieval tasks, containing over 210K fine-labeled data and 10M weak-labeled data from 51 categories and 50K+ tags with high-quality title and caption annotations.
    \item
    Based on CREATE, we introduce a novel VLM called ALWIG to address the above tasks. Our model bridges the gap between vision and language with video tags converting visual features to soft prompts and providing them to the GPT decoder for a generation. The experimental results highlight the advantages of our approach compared with other popular pre-trained models.
    \item
    We are the first to propose the task of video titling and video retrieval in the field of Chinese short videos. Our benchmark and baseline model can provide strong support for future multi-modal research and applications.
\end{itemize}

\section{Related Work}

\subsection{Benchmarks for Video-and-Language}
A large number of benchmarks have been introduced in recent years for video-and-language tasks, which cover in different filed, such as open scenario \cite{VATEX}, movies \cite{TVR}, news \cite{video-news} and e-commercial \cite{Chinese-title-pretrain}, \emph{etc}. The latest VALUE \cite{value} combines several datasets to test the performance of the model over multiple multi-modal tasks. These datasets always collect video captions annotated by a human. While these captions are valid for video captioning task, the practical applications have not yet been explored. Moreover, there are few special on Chinese corpus, except for VATEX-zh \cite{VATEX} and Poet \cite{Chinese-title-poet}. Therefore, we establish a benchmark, collect Chinese short videos, annotate high-quality annotated titles and captions for video titling and retrieval tasks.

\begin{table}
\centering
\caption{\textbf{Comparison of some relevant datasets}, the CREATE contains more open-domain videos, more annotations, and more fine-grained tag information (* indicates pre-training dataset).}
\label{tab:compare-dataset}
\resizebox{0.5\textwidth}{!}{%
\begin{tabular}{@{}c|cccccc@{}}
\toprule
\textbf{Dataset} & \textbf{Domain} & \textbf{\# Videos} & \textbf{\# Sents} & \textbf{\# Tags} & \textbf{Lang.} & \textbf{Annotation} \\ \midrule
VTW                  & Open    & 18K   & 18K  & -   & EN    & Title         \\
VATEX                & Open    & 41.3K & 826K & 600 & EN/CN & Caption       \\
BFVD/FFVD            & E-comm. & 76K   & 76K  & -   & CN    & Title         \\
T-VTD                & E-comm. & 90K   & 180K & -   & CN    & Title         \\
\textbf{CREATE210K}  & Open    & 216K  & 537K & 15,527 & CN    & Title/Caption \\ \midrule
TGIF*                & Open    & 100K  & 128K & -   & EN    & Title         \\
HowTo100M*           & Open    & 1.22M & 136M & -   & EN    & ASR           \\
WebVid-2M*           & Open    & 2.5M  & 2.5M & -   & EN    & Title         \\
Alivol-10M*          & E-comm. & 10.3M & 11M  & -   & CN    & Title         \\
\textbf{CREATE-10M*} & Open    & 10M   & 10M  & 53,044 & CN    & Title         \\ \bottomrule
\end{tabular}%
}
\end{table}


\begin{table}
    \centering
    \caption{\textbf{The splits of the whole CREATE dataset}, including 210K fine-tuning dataset and normal version 3M and large version 10M  pre-training datasets (* indicates the title is added by the user).}
    \label{tab:split-dataset}
    \resizebox{0.4\textwidth}{!}{%
    \begin{tabular}{@{}c|cccc@{}}
        \toprule
        \textbf{CREATE} & \textbf{\# Video} & \textbf{\# Title} & \textbf{\# Caption} & \textbf{\# Tag} \\ \midrule
        210K-train      & 210,493           & 210,493           & 210,493             & 15,527          \\
        210K-val        & 810               & 810$\times$10       & 810$\times$10         & 3,570           \\
        210K-test       & 5,000             & 5,000$\times$10     & 5,000$\times$10       & 1,191           \\
        3M-pretrain     & 3M                & 3M*               & -                   & 45,277          \\
        10M-pretrain    & 10M               & 10M*              & -                   & 53,044          \\ \bottomrule
    \end{tabular}%
    }
\end{table}

\begin{table*}[]
\centering
\caption{\textbf{The performance of proposed simple two-steam video-text matching scorer with a variety of backbones.} We try to pre-train these models under different corpus end-to-end. The experimental results on the VATEX public-test show that using the ViT-BERT model pre-trained on the video-title of Chinese short videos can help to learn better alignment efficiently, which is used to filter out bad videos for CREATE datasets.}
\resizebox{0.8\textwidth}{!}{%
\begin{tabular}{@{}ccccccclccc@{}}
\toprule
\multirow{2}{*}{\textbf{\#}} & \multirow{2}{*}{\textbf{Model}} & \multirow{2}{*}{\textbf{Pre-trained Dataset}} & \multirow{2}{*}{\textbf{Finetune}} & \multicolumn{3}{c}{\textbf{Text-Video Retrieval}} &  & \multicolumn{3}{c}{\textbf{Video-Text Retrieval}} \\ \cmidrule(lr){5-7} \cmidrule(l){9-11} 
 &  &  &  & \multicolumn{1}{l}{R@1} & \multicolumn{1}{l}{R@5} & \multicolumn{1}{l}{R@10} &  & \multicolumn{1}{l}{R@1} & \multicolumn{1}{l}{R@5} & \multicolumn{1}{l}{R@10} \\ \midrule
1 & \multirow{5}{*}{S3DG-BERT} & - & $\checkmark$ & 1.7 & 5.5 & 8.6 &  & 5.8 & 15.4 & 21.7 \\
2 &  & HowTo100M-CN & $\times$ & 4.2 & 13.2 & 19.3 &  & 4.4 & 16.1 & 23.7 \\
3 &  & HowTo100M-CN & $\checkmark$ & 7.7 & 21.8 & 30.5 &  & 15.4 & 39.6 & 53.9 \\
4 &  & Random-10M & $\times$ & 6.8 & 20.1 & 29.5 &  & 11.8 & 30.1 & 41.9 \\
5 &  & Random-10M & $\checkmark$ & 19.6 & 47.7 & 61.7 &  & 31.2 & 60.4 & 72.1 \\ \midrule
6 & \multirow{2}{*}{\begin{tabular}[c]{@{}c@{}}TimeSformer\\ -BERT\end{tabular}} & Random-10M & $\times$ & 15.3 & 37.5 & 49.3 &  & 31.7 & 62.0 & 74.3 \\
7 &  & Random-10M & $\checkmark$ & \textbf{43.1} & \textbf{77.1} & \textbf{86.9} &  & 60.8 & 88.0 & 94.0 \\ \midrule
8 & \multirow{2}{*}{ViT-BERT} & Random-10M & $\times$ & 18.8 & 44.0 & 56.4 &  & 37.2 & 66.6 & 77.5 \\
9 &  & Random-10M & $\checkmark$ & 41.1 & 75.3 & 85.2 &  & \textbf{64.2} & \textbf{89.6} & \textbf{94.3} \\ \bottomrule
\end{tabular}%
}
\label{tab:two-stream-result}
\end{table*}

\subsection{Video-and-Language Pre-training}
Thanks to some large-scale datasets, such as HowTo100M \cite{HowTo100M} and WebVid \cite{frozen}, downstream VL tasks can be greatly improved by weakly supervised learning through narration-video or title-video pairs. 
According to the main structure of the model, the pre-trained model can be divided into one-stream and two-stream. One-stream models \cite{HERO, ActBERT} always design various proxy tasks, fusion multi-modals through a transformer-based model, and adapt to discriminative or generative tasks simultaneously. Two-stream models \cite{MIL-NCE,clip4clip, frozen} leverage two separate backbones and contrastive learning to align visual and text. Our model combines the advantages of both to take alignment efficiently through the two-stream model and to generate accurately through the one-stream model. 

\subsection{Video Titling and Video Retrieval Tasks}
The earliest work is VTW \cite{VTW}, which collects video titles in the wild and combines highlight detection and title at the same time. However, the performance is not satisfactory due to the amount of data and the lack of pre-training techniques. The most relevant works are \cite{Chinese-title,Chinese-title-poet,Chinese-title-pretrain}. They collect videos from Taobao and annotate titles in Chinese. Although good results have been achieved, the videos are limited to the e-commerce field. For the retrieval task, while some refined approaches \cite{Object-Relation, open-book} have been developed and remarkable progress has been made, there are still limitations in the efficiency for practical application.

\section{CREATE Benchmark}

\subsection{Dataset Collection}
As mentioned above, the bottleneck of the practical application of Chinese short video titling and retrieval is the lack of appropriate dataset. In addition, it is effective to improve the performance of downstream tasks through pre-training without increasing annotations. Therefore, we construct a high-quality labeled dataset CREATE-210K and two large-scale weak-labeled datasets CREATE-3M/10M from the Tencent video platform.

\textbf{High-quality Labeled CREATE-210K}.
We start by building a video tagging system that contains 51 categories and over 50K video tags for a wide coverage of short video content. Each video always has one category and several tags, which can be seen as the coarse-grained and fine-grained classifications of the video. Then we follow the principle of collecting at least five videos per tag to ensure enough data for training in different domains. We do not average sample videos according to the video categories since some categories, \emph{e.g.}, military and financial, are not common. Filtered by video tagging system, we try to avoid including some special tags, \emph{e.g.}, film or television variety shows, since understanding these videos requires lots of additional information, such as stars and plots, which is difficult to annotate and model. The distributions of tags and categories are illustrate in Figure \ref{fig:tag-cate-distribution}. Sorted by the number of video categories and tags, the video contents mainly focus on ``\textit{people daily life}'', ``\textit{instruction video}'' and ``\textit{animals show}'', \emph{etc}. The long-tail problem is inevitable because tags are hierarchical, and some tags are subsets of larger concept tags, \emph{e.g.}, ``\textit{monk parrot}'' belongs to ``\textit{cute animals}''. 

We limit the video time to less than 60 seconds and eventually collected over 210K short videos for next step annotations.More than 100 workers are involved in annotation tasks to ensure data diversity. To obtain high-quality annotations, each worker has undergone rigorous training and testing, with a clear definition of the difference between video titling and captioning. Moreover, workers are provided categories and tags of each as hints and required to use information as much as possible. A word such as ``\textit{things}'' should be replaced with specific objects since this could increase diversity and avoid general annotation. The word limits for video titles and captions are 15$\sim$30 and 25$\sim$50, respectively. Finally, it takes half a year to collect and check more than 537k annotations. More details about annotation rules and interface are shown in the appendix.

\begin{figure*}
    \centering
    \includegraphics[width=1\linewidth]{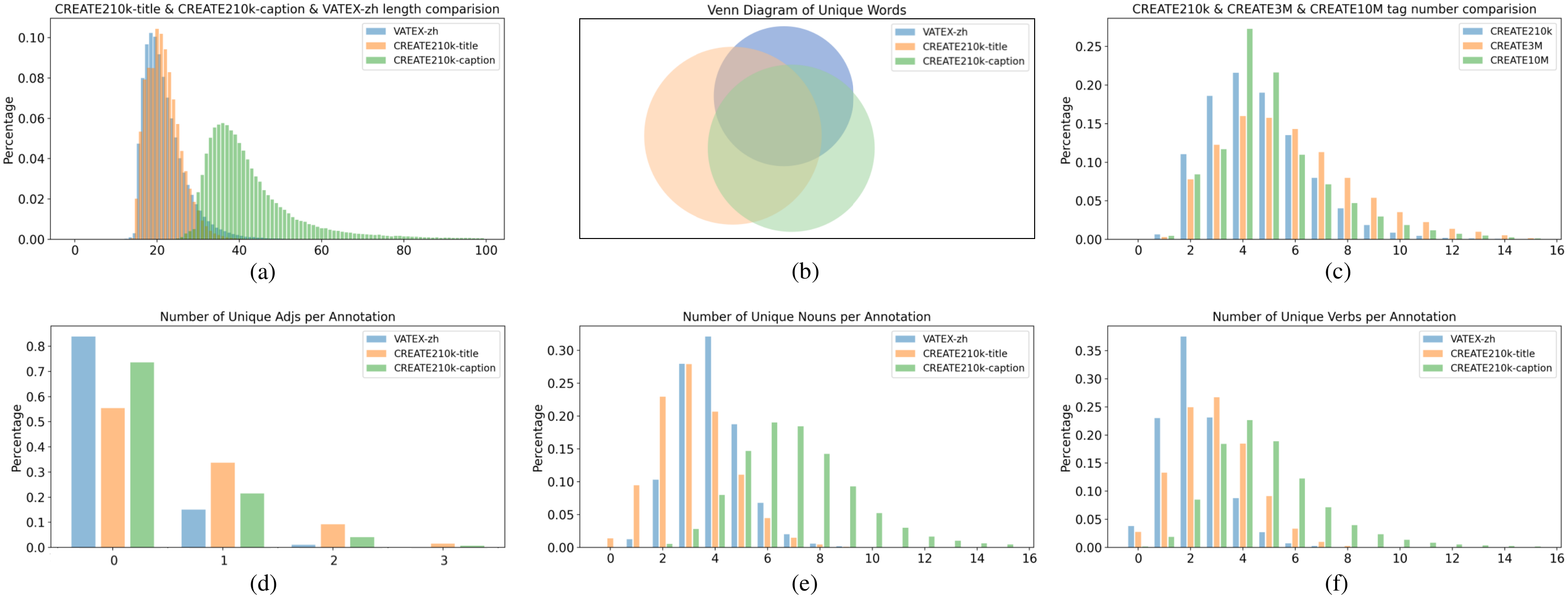}
    \caption{\textbf{Some statistics on the datasets} indicate our datasets and annotations have better diversity. (a) indicates the distribution of the annotation length in three datasets. (b) indicates the inclusion relation of unique words by Venn Diagram in three datasets. (c) represents the distribution of the number of tags. (d)-(f) shows the distribution on three part-of-speech of unique words.}
    \label{fig:statistics-dataset}
    
\end{figure*}

\textbf{Large-scale Weak-labeled CREATE-3M/10M}.
To increase the generalization of the model and the extensibility of the tasks, we establish two large-scale weak-labeled datasets CREATE-3M/10M. Each video has its category, several tags and an original title. The noise within videos and titles is the most serious problem, \emph{e.g.}, some videos or titles are of poor quality, and some video-title pairs are mismatched. 
In order to filter out these low-quality videos, we designed an efficient automatic filter to determine the consistency of videos and titles inspired by the work \cite{MIL-NCE,Timesformer,clip4clip}.
As shown in Figure \ref{fig:baseline-structure-compare}.a, a two-stream model leverages a visual encoder and a  textual encoder to extract visual and textual features separately in an end-to-end manner. The contrastive learning is leveraged to push mismatched and pull matched features, achieving alignment between both modalities.  

We conduct comparison experiments on three backbones, \emph{i.e.}, S3DG-BERT, TimeSformer-BERT and ViT-BERT, to verify which is the most suitable backbones for feature extraction, as illustrated in Table \ref{tab:two-stream-result}. We first evaluate the S3DG-BERT on the translated HowTo100M and randomly collected 10M video-title datasets following the same setting as the previous work \footnote{https://github.com/antoine77340/MIL-NCE\_HowTo100M}. It shows that learning with video-title pairs can achieve better alignment than video-narrations in instructional videos\footnote{Note: This does not exclude the reason for the increased noise introduced by translation.}, as shown in Table \ref{tab:two-stream-result}.Line2,4. Besides, we evaluate the transformer-based models, \emph{i.e.} TimeSformer-BERT and ViT-BERT. Compared with 3D-CNN based model, the transformer-base model can better support large-scale data during pre-training, as shown in Table \ref{tab:two-stream-result},Line4,6,8. Moreover, compared with spatial and temporal attentions in TimeSformer, we conduct average pooling in the temporal dimension over 8 frames. Although the performance is slightly reduced, it is more efficient for calculating matching scores and extracting visual features. Eventually, we choose ViT-BERT model pre-trained on random-10M videos as the video-title matching scorer. Videos with matching scores of less than 0.3 are filtered out, and most of the remaining videos form the final pre-trained dataset, \emph{i.e.}, the normal version CREATE-3M and large version CREATE-10M. The normal version is more convenient for algorithm iteration.

\begin{figure*}
    \centering
    \includegraphics[width=1\linewidth]{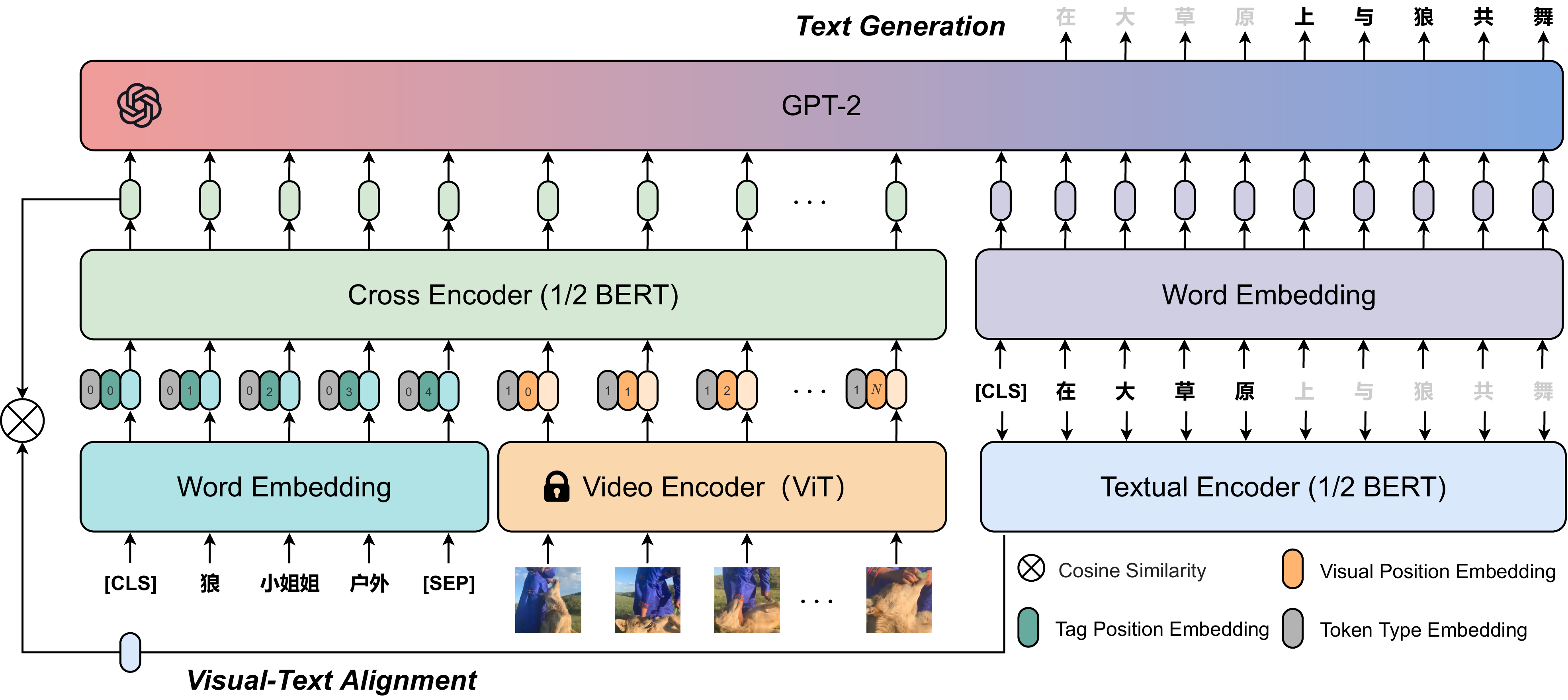}
    \caption{\textbf{Overall framework of our proposed ALWIG model.} ALWIG consists of a tag-driven video-text alignment module and a GPT-based generation module for video titling and retrieval tasks.}
    \label{fig:alwig_framework}
\end{figure*}

\subsection{Statistics of the CREATE dataset}
We analyse the CREATE dataset in terms of video information and annotations. As illustrated in Table \ref{tab:split-dataset}, we collected a total of 210,493 videos for training, with one title and caption annotated, 810 videos for validation, and 5,000 videos for testing, each video has 10 titles and captions. For the weak-labeled dataset, we filter out 3M and 10M videos with their original titles. The average video length is around 30 seconds. The distributions of annotation length are illustrated in Figure \ref{fig:statistics-dataset}.a. The average title or caption lengths of VATEX-zh, CREATE210K-caption, CREATE210K-title, CREATE3M and CREATE10M are 22.45, 43.54, 21.71, 20.03 and 23.96. The distributions of tag numbers are illustrated in Figure \ref{fig:statistics-dataset}.c. The average number of tags within CREATE210K, CREATE3M and CREATE10M are 4.65, 5.73 and 5.03.  

In addition to basic information, we pay more attention to the richness of content covered by the annotations. We use the Venn Diagram to depict the approximate inclusion relation of unique words in VATEX-zh, CREATE210k-title and CREATE210k-caption, as shown in Figure \ref{fig:statistics-dataset}.b. Our CREATE210k-caption covers 72.85\% of the vocabulary of VATEX-zh, and 62.68\% of the unique words do not appear in VATEX-zh. Moreover, we analyze parts of speech(POS) of annotations, \emph{i.e.}, of the above three datasets. As shown in  Figure \ref{fig:statistics-dataset}.d-f, as can be seen from the distributions of the three POS, our datasets contain more information in each annotation. 


\section{ALWIG Method}

ALWIG consists of a tag-driven video-text alignment module and a GPT-based generation module for video titling and retrieval tasks, as shown in Figure \ref{fig:alwig_framework}.
We use a 12-layer transformers ViT-B/16 as the video feature extractor, and initialized it with the weights from CLIP model \footnote{https://github.com/openai/CLIP}. The video clip is encoded into a sequence of $N$ video features $V=\{\bm{v}_1, \cdots, \bm{v}_N\}$. We get $M$ tag embeddings $O= \{\bm{o}_{\rm{cls}}, \bm{o}_1, \cdots, \bm{o}_M, \bm{o}_{\rm{sep}}\}$ using the word embedding in BERT.

\textbf{Tag-driven video-text alignment module}.
We concatenate tag embeddings and video features into $\{\bm{o}_{\rm{cls}}, \bm{o}_1, \cdots, \bm{o}_M, \bm{o}_{\rm{sep}}, \bm{v}_1, \cdots, \bm{v}_N\}$, where $\bm{o}_{\rm{cls}}$ and $\bm{o}_{\rm{sep}}$ are embeddings of $[\rm{CLS}]$ and $[\rm{SEP}]$ tokens. 
Furthermore, we use two independent 6-layer transformers as the cross-encoder and the textual-encoder. Both encoders are initialized with the first six layers transformer of Bert model \footnote{https://huggingface.co/bert-base-chinese}. 
The cross-encoder is utilized to integrate video features and tag embeddings into fusion embeddings $F=\{\bm{f}_{\rm{cls}}, \cdots, \bm{f}_{\rm{M+N}}\}$, where $\bm{f}_{\rm{cls}}$ can be regarded as the fused video-tag representation driven by the tags. 
Textual encoder embeds a text input $T$ into a sequence of $L$ token embeddings $W=\{\bm{w}_{\rm{cls}}, \bm{w}_1, \cdots, \bm{w}_L\}$, where $\bm{w}_{\rm{cls}}$ represents the whole representation of textual embedding.
The video-text alignment is to learn a similar function $s(F , W) = \phi(\bm{f}_{\rm{cls}})^{\rm{T}} \psi(\bm{w}_{\rm{cls}})$ between visual and textual embeddings via contrastive learning, where $\phi(\cdot)$ and $\psi(\cdot)$ are the linear functions with normalization to map each embedding into the common semantic space. We follow the infoNCE loss function as shown in Equation \ref{eq2}, where $\tau$ is a learnable temperature coefficient, and $W_+$ and $F_+$ are positive samples within batch.

\begin{equation}
\label{eq2}
\begin{split}
\mathcal{L}_{align}=-\log \frac{\exp \left(s(F,W_{+}) / \tau\right)}{\sum_{i=1}^{K} \exp \left(s(F,W_{i}) / \tau\right)} \\
-\log \frac{\exp \left(s(W,F_{+}) / \tau\right)}{\sum_{i=1}^{K} \exp \left(s(W,F_{i}) / \tau\right)}.
\end{split}
\end{equation}

\begin{figure*}
    \centering
    \includegraphics[width=\linewidth]{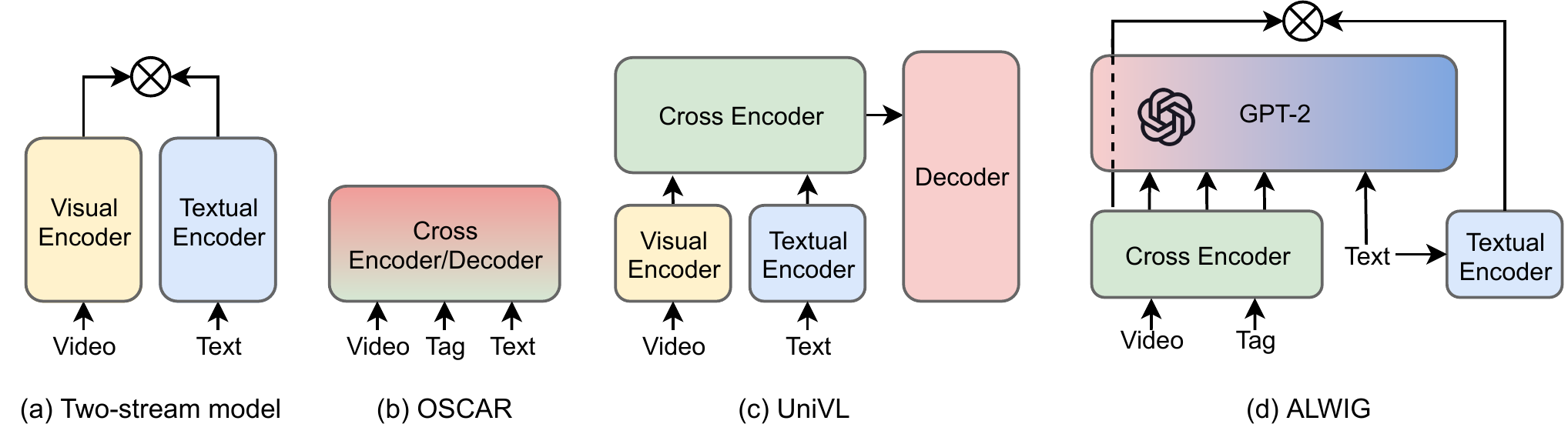}
    \caption{\textbf{The mean structure of the models.} (a) indicates the two-stream model, which we used as a video-text matching scorer for filtering the 10M pre-training dataset. The model is trained in an end-to-end manner. (b) shows the OSCAR model, which is a simple cross encoder/decoder structure. We replace the object features with frames features as input. (c) shows the UniVL model, a typical encoder and decoder structure, and it is a popular video-text pre-trained model. (d) indicates our ALWIG model, which leverages the tag-driven fusion via contrastive learning to achieve alignment, and the power of GPT-2 to achieve generation. }
    \label{fig:baseline-structure-compare}
\end{figure*}

\textbf{GPT-based generation module}.
One of the downstream tasks we are most interested in is video titling. It requires to express not only the general meaning of the video but also subjective expressions to attract the audiences' interests. 
Therefore, we leverage the power of GPT as the decoder to help introduce external linguistic knowledge to reduce the difficulty of textual generation and improve the generalization of the model \cite{Object-Relation}. 
The input of the decoder is the fusion embeddings $F$ mentioned above, and the output is the ground-truth text. We utilize the typical auto-regressive training method to train the model following cross-entropy loss as shown in Equation \ref{eq1}, where $F$ can be seen as a soft-prompts for the generation.

\begin{equation}
\label{eq1}
\mathcal{L}_{gen}=- \sum_{l=1}^{L} \log p_{\theta}(T_l|T_{<l},F).
\end{equation}

In summary, we utilize a tag-driven cross-encoder with the help of contrastive learning to align the modalities and take advantage of a GPT-based decoder to generate text. The full pre-training objective of ALWIG is:

\begin{equation}
\mathcal{L}=\mathcal{L}_{align} + \mathcal{L}_{gen}.
\end{equation}

\begin{table*}[]
\centering
\caption{The experimental results of UniVL, OSCAR and proposed ALWIG. All the models are pre-trained on two large-scale weak-labeled dataset (3M=CREATE3M, 10M=CREATE10M) and fine-labeled on CREATE210K for three tasks: video retrieval, titling and captioning. The bottom three lines represent the ablation studies. }
\label{tab:baseline-result}
\resizebox{\textwidth}{!}{%
\begin{tabular}{@{}ccccccccccccc@{}}
\toprule
\multirow{2}{*}{\textbf{\#}} & \multirow{2}{*}{\textbf{Model}} & \multirow{2}{*}{\textbf{\begin{tabular}[c]{@{}c@{}}Pre-trained\\ Dataset\end{tabular}}} & \multicolumn{2}{c}{\textbf{Task1: Video Retrieval}} & \textbf{} & \multicolumn{3}{c}{\textbf{Task2: Video Titling}} & \textbf{} & \multicolumn{3}{c}{\textbf{Task3: Video Captioning}} \\ \cmidrule(l){4-13} 
 &  &  & \textbf{T2V Recall@1/5/10} & \textbf{V2T Recall@1/5/10} & \textbf{} & \textbf{CIDEr} & \textbf{BLEU-4} & \textbf{Rouge-L} & \textbf{} & \textbf{CIDEr} & \textbf{BLEU-4} & \textbf{Rouge-L} \\ \midrule
1 & \multirow{2}{*}{UniVL} & 3M & 59.3 / 83.9 / 90.4 & 73.6 / 90.7 / 94.8 & & 13.3 & 6.4 & 26.1 & & 18.9 & 14.0 & 33.2  \\
2 &  & 10M & 61.7 /85.1 / 91.3 & 76.8 / 92.1 / 95.9 &  & 13.8 & 6.9 & 26.5 &  & 22.9 & 14.5 & 33.4 \\ \midrule
3 & \multirow{2}{*}{OSCAR} & 3M & 61.3 / 84.6 / 90.7 & 74.9 / 91.5 / 95.1 &  & 35.8 & 8.4 & 29.8 &  & 34.7 & 14.2 & 33.4 \\
4 &  & 10M & 62.1 / 85.5 / 91.3 & 75.2 / 91.5 / 95.5 &  & \textbf{36.3} & 9.0 & 30.7 &  & 35.2 & 15.2 & 33.8 \\ \midrule
5 & \multirow{2}{*}{ALWIG} & 3M & 60.7 / 85.0 / 91.0 & 75.3 / 91.9 / 96.0 &  & 35.5 & 9.3 & 31.0 &  & 32.2 & 14.9 & 34.6 \\
6 &  & 10M & \textbf{65.6 / 87.7 / 92.7} & \textbf{79.3 / 93.9 / 96.8} &  & 36.1 & \textbf{9.7} & \textbf{31.6} &  & \textbf{35.9} & \textbf{16.3} & \textbf{35.5} \\ \midrule
 & \multicolumn{2}{l}{\textbf{Ablation Study}} &  &  &  &  &  &  &  &  &  &  \\ \midrule
7 & Baseline & 3M & \textbf{60.7 / 85.0 / 91.0} & \textbf{75.3 / 91.9 / 96.0} &  & \textbf{35.5} & \textbf{9.3} & \textbf{31.0} &  & \textbf{32.2} & \textbf{14.9} & \textbf{34.6}  \\
8 & w/o Tag & 3M & 51.7 / 79.0 / 86.8 & 67.1 / 87.5 / 92.7 &  & 15.6 & 6.5 & 27.2 &  & 13.9 & 12.0 & 31.9 \\
9 & w/o GPT & 3M & 58.7 / 83.5 / 90.1 & 72.2 / 90.2 / 94.8 &  & 26.7 & 6.6 & 28.5 &  & 29.1 & 12.8 & 34.3 \\ 
10 & w/o pretrain & - & 43.0 / 72.3 / 81.8 & 56.6 / 80.9 / 88.2 &  & 21.4 & 6.1 & 28.1 &  & 23.8 & 13.9 & 33.7 \\ \bottomrule
\end{tabular}%
}
\end{table*}

\section{Experiments}

\subsection{Baseline Models}
We benchmark three representative vision and language pre-trained models, \emph{i.e.}, OSCAR and UniVL, and our ALWIG on the proposed CREATE dataset. Both models are pre-trained on the large-scale pre-training dataset through multiple proxy tasks, such as mask tokens prediction, contrastive learning on visual-textual pairs, \emph{etc}, then finetuned on many downstream tasks, such as cross-modal retrieval, captioning, VQA, \emph{etc}. In the model structure, both models adopt the general transformer-based structure.

As shown in Figure \ref{fig:baseline-structure-compare}.b, OSCAR leverages an integrated encoder-decoder to fuse visual and textual features from beginning to end. It controls generation or discriminative tasks by setting different types of masks. Moreover, the most instructive thing in OSCAR is using object tags as anchor points to align the image and language modalities in a shared semantic space. 

Instead, UniVL is a flexible model for most of the multimodal downstream tasks considering both efficiency and effectiveness, as illustrated in Figure \ref{fig:baseline-structure-compare}.c. It utilizes two independent encoders to enhance the representation of each modal at the beginning and leverages cross-encoder to fuse each other. Besides, a separate encoder-decoder can explicitly handle generation tasks, which is more flexible.

\subsection{Experimental Setting}
Our model consists of two half $\rm{BERT}_{base}$ with 123.7M parameters and a $\rm{GPT}_{base}$ with 154.5M parameters. We pre-train the model for 30 epochs using a batch size of 32 on 48 NVIDIA A100 GPUs. We use the AdamW optimizer with a weight decay of 0.02. The learning rate is warmed-up to $1e^{-5}$ in the first 10 epochs and decayed to $1e^{-6}$ following a cosine schedule. Before training, we extract video features 1fps via ViT pre-trained by CLIP. The dimension of the hidden state in BERT and GPT is 768, and the output features of the two-stream structure are mapped to 512. We utilize beam-search (beam size=3) for the generation.

We leverage the standard captioning metrics, \emph{i.e}, BLEU-4 \cite{Bleu}, CIDEr \cite{CIDEr} and Rouge-L \cite{Rouge}, to measure the performance of video titling and captioning. While the rules-based metrics seem unable to reflect the quality of the video titling model for findability and clickability, the 10-annotations per video alleviates this problem to some extent, and the learnable metrics are left for the future.
We split the words using Jieba Chinese text segmentation \footnote{https://github.com/fxsjy/jieba}, and we count 4 continuous words as 4-grams. It can be noted that we are not using Meteor as the metric since Meteor \cite{Meteor} defaults to calculate the relationship of English synonyms but does not contain the Chinese thesaurus. 
Furthermore, we utilize metrics Recall at K (Recall@K) to measure video-text retrieval performance. R@K measures the proportion of correct targets retrieved from K samples.

\subsection{Results and Analysis}
We adapt three pre-trained models \emph{i.e.}, UniVL, OSCAR and ALWIG to three tasks, as shown in \ref{tab:baseline-result}. Three models are all pre-trained on CREATE3M/10M with their best performances. Compared with UniVL, our model has significantly improved in all three tasks, especially the performance of ALWIG improve by 62.5\% on CIDEr in the video titling task, indicating that many novel words have been generated, which is attributed to the use of tags through tag-driven fusion module. 
Compared to the automatically detected tags in OSCAR, the manually collected tags in our work are more rich and high-quality. Instead of focusing on how to get these tags, we try to use them directly as a bridge for the alignment between multiple modalities in pre-training or external knowledge for downstream tasks. Compared with OSCAR on retrieval task, our model is superior, which also illustrates the validity of the GPT model

In addition, to illustrate the effectiveness of our proposed two modules in more detail, we conducted multiple ablation experiments. 
When the tag-driven fusion module is removed, the performance of CIDEr is significantly reduced by 56\% and 56.8\% respectively on the video titling and captioning tasks, and 14.8\% reduces Recall@1 on the text-video retrieval task. It is worth noting that the decline of other metrics in the video description task is not particularly severe, which indicates that it has a small impact on the overall sentence pattern and also reflects that the description task pays more attention to the main content of the video rather than some novel expression. 
When the GPT module is removed, the retrieval performance slightly decreases, while the BLEU-4 of the two generation tasks decreases by 34.4\% and 14.1\% respectively, indicating that GPT is of great help to the learning of basic sentence patterns and can reduce the difficulty of generation. 
Meanwhile, we also experiment with the model without pre-training, and all indicators are far lower than the pre-training model's, indicating that the pre-training and fine-tuning paradigm can effectively improve the model's performance.

\section{Conclusion}
In this paper, we establish the first Chinese short video retrieval and title generation benchmark, \emph{i.e.}, CREATE to facilitate research and application for video retrieval and titling tasks. The CREATE contains a high-quality fine-labeled dataset and two large-scale pre-training datasets.  A large number of statistics indicate that our datasets have richer visual content and annotations. Based on CREATE, we propose a novel model ALWIG to better accomplish the video retrieval and generation tasks with the power of tag-driven fusion and the GPT model. Extensive experiments verify the validity of our model and provide some good baselines for future research.

\bibliography{reference}
\bibliographystyle{acl_natbib}

\newpage
\appendix
\label{sec:appendix}
\section{Annotation Details}
In this section, we introduce our annotation details for CREATE-210K to deepen the user's understanding of the dataset. This introduction is in the actual labeling process requires the workers to read in advance, and after a multi-questions test to meet the standards to enter the formal labeling link.

\subsection{Video Captioning Details}

\begin{enumerate}
    \item \textbf{About the degree of labeling.} 
    Highlight the key elements of the video, such as the key event, people and objects. The general description should be avoided. 
    
    \item \textbf{About the objectivity.} 
    Describe the content of the video objectively without mixing any personal emotions and comments, such as ``\textit{that's awesome}'', ``\textit{it's really}'', ``\textit{it/that looks}''.
    
    \item \textbf{About the source of information.} 
    Try to describe what you see. Narration, subtitles, and conversations are only used as supporting information. Do not describe background music.
    
    \item \textbf{About the usage of tags.} 
    Use the tags provided as much as possible. Refer to the tags for some objects if you don't recognize them. Avoid general meaningless expressions such as ``\textit{something}'', ``\textit{object}'', ``\textit{liquid}''. Fine-grained tags of objects should be used, \emph{e.g.}, tags provide ``\textit{agates}'' that cannot simply be described as ``\textit{stone}''. 
    
    \item \textbf{About stars and varieties.}
    Should be specifically stated the name of the person and the name of the variety, if the label has relevant tips.
    
    \item \textbf{About wearing description.}
     If wearing is not the focus of the video please do not describe wearing throughout, while avoiding the use of templates, such as a large number of the expressions with ``\textit{A person wearing}'', will be considered invalid labels.
     
    \item \textbf{Describe directly.}
    The descriptions like ``\textit{I/We can see}'', ``\textit{the video is the description of}'', ``\textit{a person is facing the camera/phone}'' are not available.

    \item \textbf{About hands.} In some videos, there is no person but a pair of hands instead, such as cooking, please infer the gender or occupation of the person, do not say what one hand is doing.

\end{enumerate}

\subsection{Video Titling Details}

\begin{enumerate}
    
    \item \textbf{Reject the clickbaits.}
     Avoid exaggerate content, such as ``\textit{Shock!}'',  no actual content ``\textit{LOL}'', and adult content.
     
    \item \textbf{About attraction.} Attraction needs express interesting content through the title, not touts, such as ``\textit{Take a look at it!}''

    
    \item \textbf{About plagiarism.} The title can not be copied, must be original. The use of video audio or subtitles in some words is acceptable, but not more than 80\% of the original video content.
    
\end{enumerate}

\begin{figure}
    \centering
    \includegraphics[width=1\linewidth]{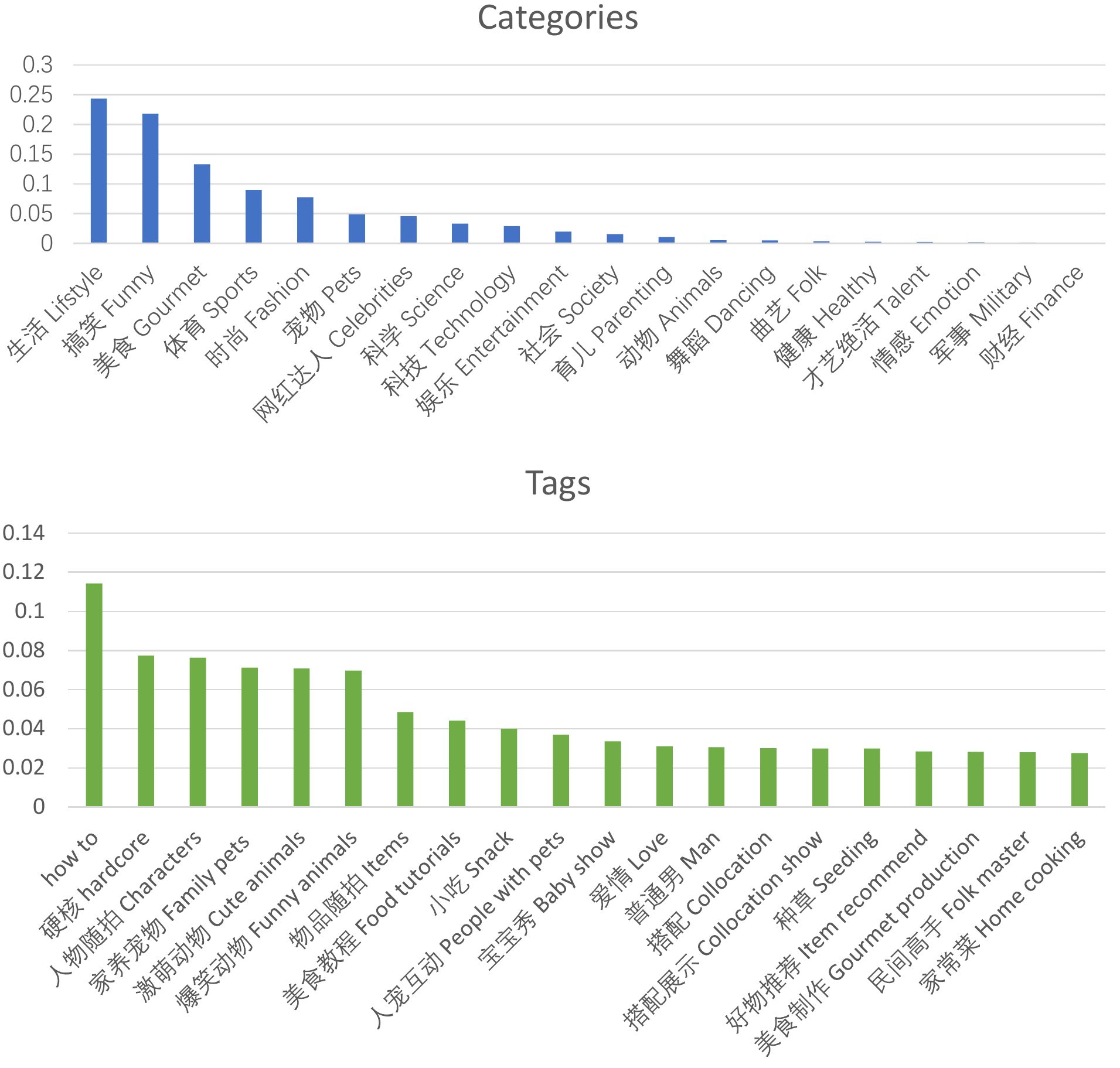}
    \caption{The distributions of categories and tags in the CREATE dataset.}
    \label{fig:tag-cate-distribution}
\end{figure}

\subsection{The interface of annotation}
In this section, we demonstrate the interface during the labelling process. Workers are provided with the video content, original video title and video tags to help to label. At the same time as labelling, it is necessary to check which tags are used to force workers to use them as much as possible, thus improving the quality of labelling.

\begin{figure*}
    \centering
    \includegraphics[width=1\linewidth]{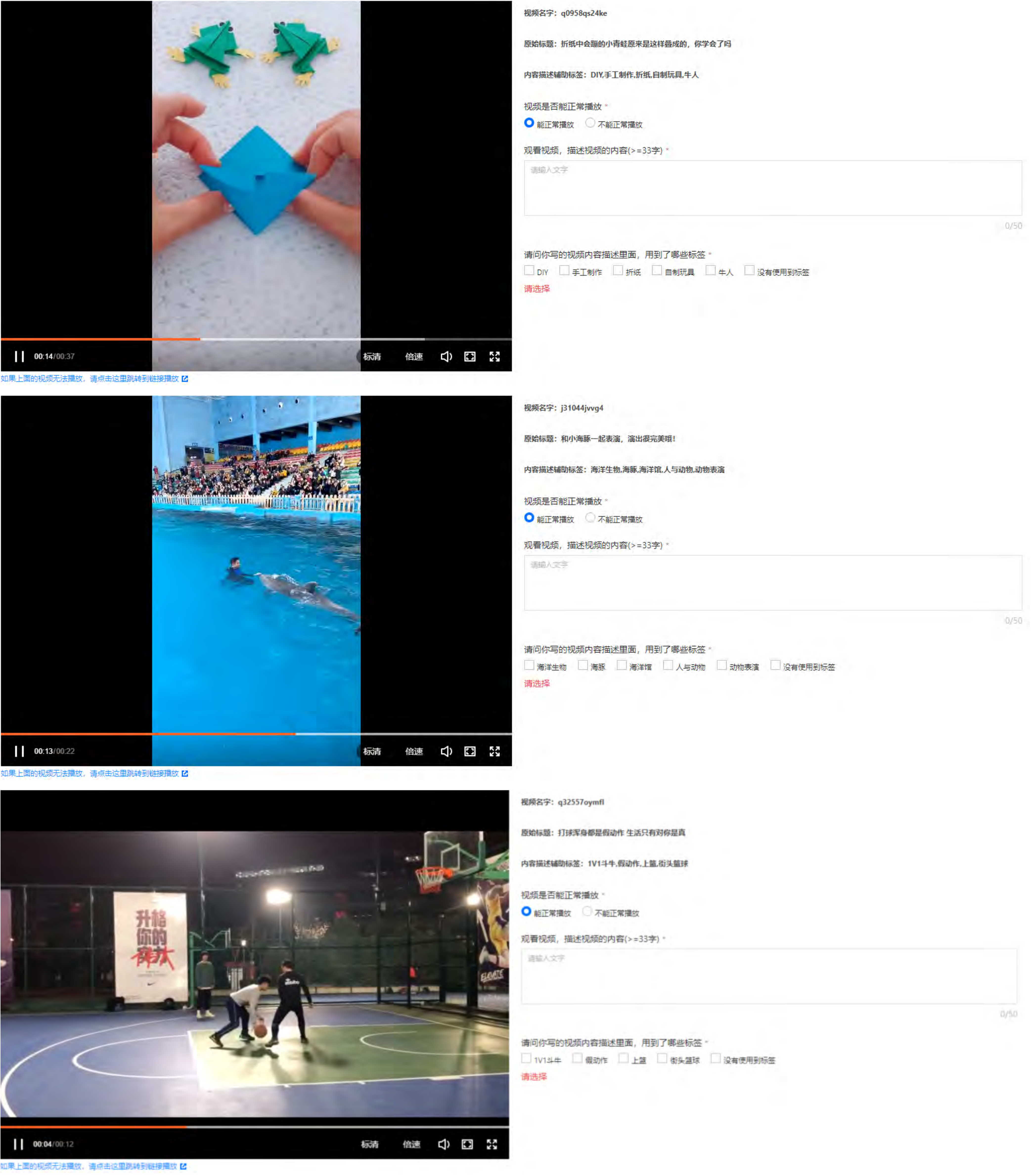}
    \caption{The demonstration of the interface during the actual labeling process.}
    \label{fig:interface}
\end{figure*}

\section{Result Details}
\begin{figure}
    \centering
    \includegraphics[width=\linewidth]{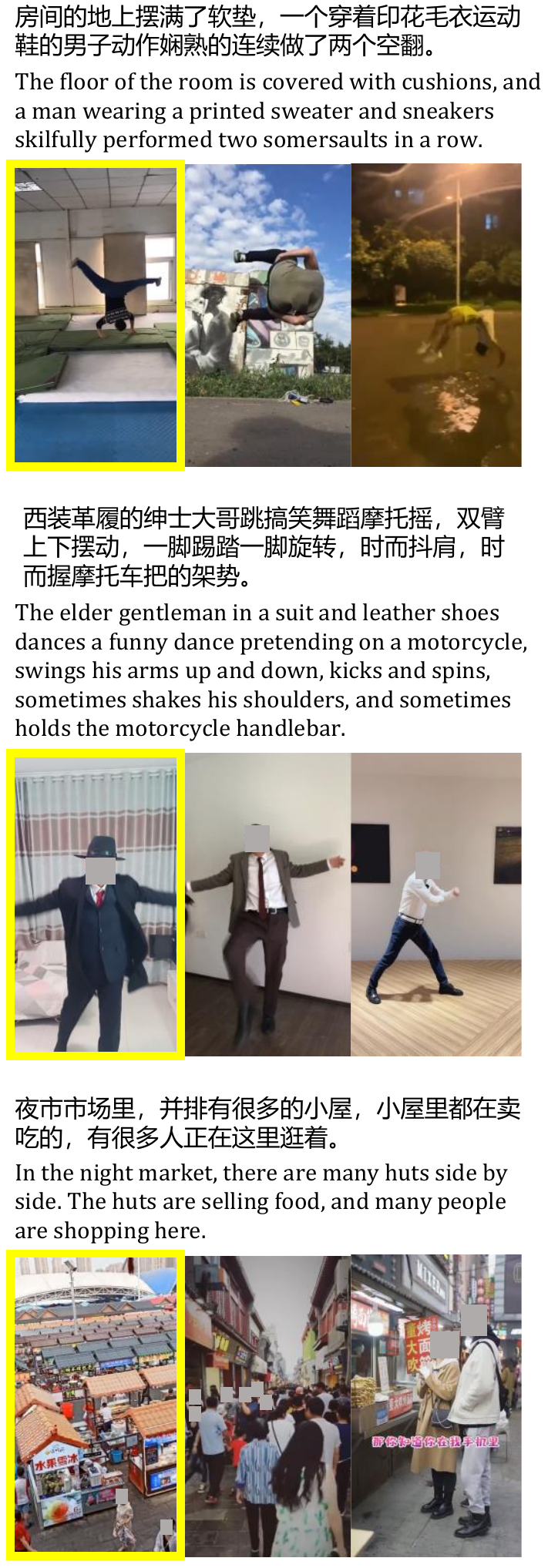}
    \caption{The visualization of text-to-video retrieval.}
    \label{fig:t2v}
\end{figure}

\begin{figure}
    \centering
    \includegraphics[width=0.8\linewidth]{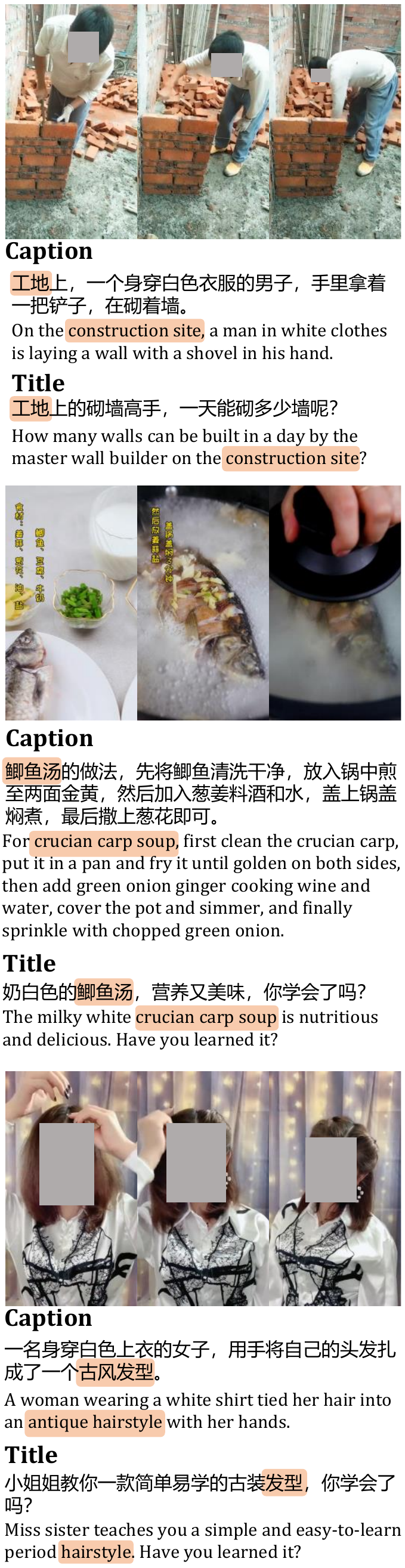}
    \caption{The generated video captions and titles.}
    \label{fig:cap_tit_gen}
\end{figure}

\begin{figure*}
    \centering
    \includegraphics[width=1\linewidth]{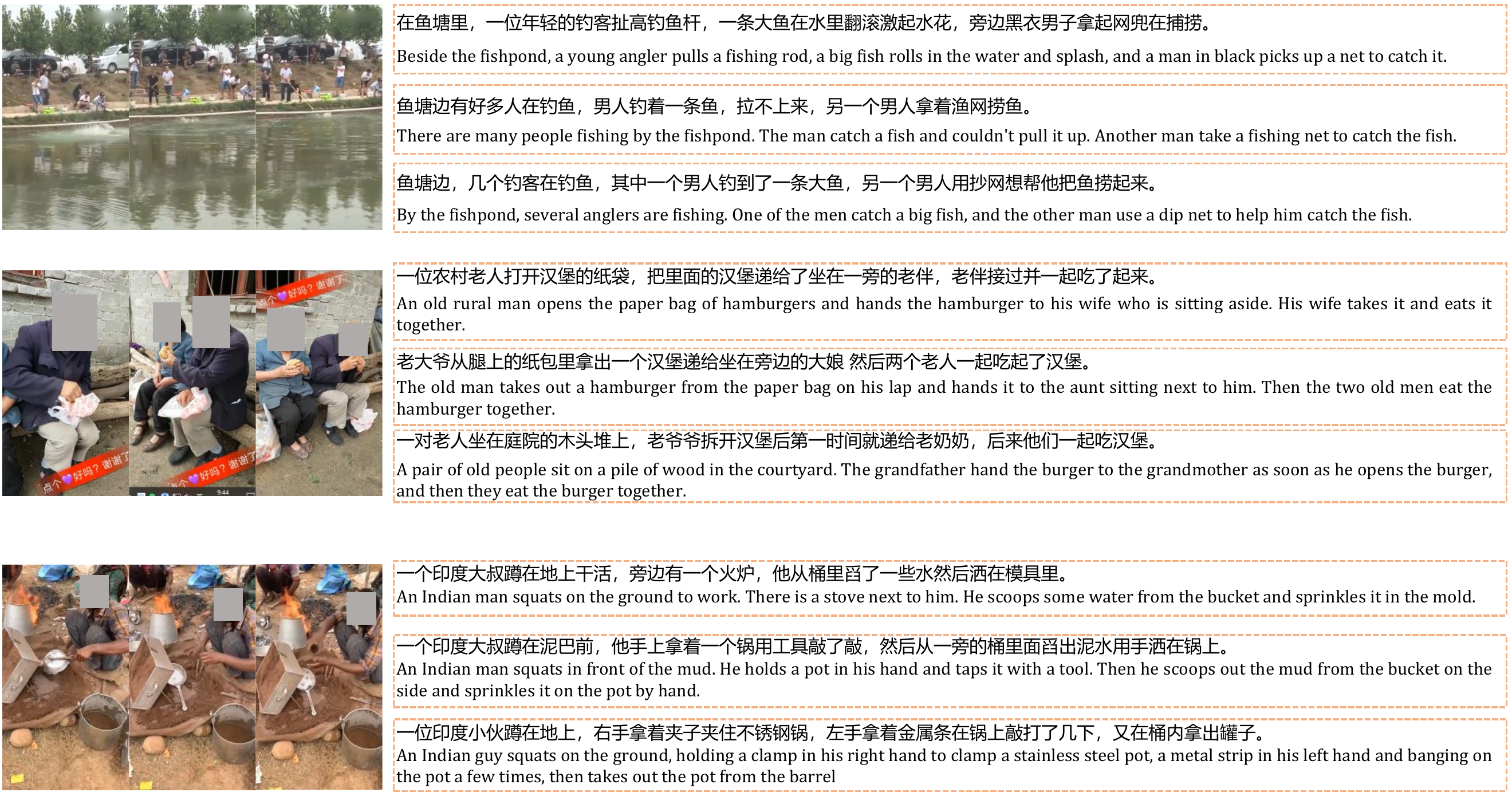}
    \caption{The visualization of video-to-text retrieval.}
    \label{fig:v2t}
\end{figure*}

In this section, we demonstrate some cases sampled from the result on video retrieval, titling and captioning tasks, as shown in Figure \ref{fig:t2v}, \ref{fig:cap_tit_gen} and \ref{fig:v2t}.

\end{document}